\documentclass[12pt, copyright, goog]{google}
\usepackage{microtype}
\usepackage{graphicx}
\usepackage{amsmath}
\usepackage{graphicx}
\usepackage{amsmath}
\usepackage{hyperref}
\usepackage{url}
\usepackage{booktabs}
\usepackage{multirow}
\usepackage{siunitx}
\usepackage{enumitem}
\usepackage{subcaption}
\usepackage{wrapfig}
\usepackage[capitalize,noabbrev]{cleveref}
\usepackage[table]{xcolor}
\usepackage{tcolorbox}
\tcbuselibrary{skins, breakable}
\usepackage{lineno}

\definecolor{darkblue}{rgb}{0, 0, 0.5}
\hypersetup{colorlinks=true, citecolor=darkblue, linkcolor=darkblue, urlcolor=darkblue}

\usepackage[authoryear, sort&compress, round]{natbib}
\uselogo{} 

\title{LASER: Language Model Regression for Semi-Structured Workflow Resource and Runtime Estimation}

\correspondingauthor{my-email@google.com}

\reportnumber{0001} % Leave blank if n/a

\correspondingauthor{shengkezhou@google.com, \{y\_yin, lip\}@ucsb.edu}

\reportnumber{ } % Leave blank if n/a

\renewcommand{\today}{}

\author[$^\dagger$,1, 2]{Yuxuan Yin}
\author[1]{Shengke Zhou}
\author[1]{Yunjie Zhang}
\author[1]{Ajay Mohindra}
\author[2]{Boxun Xu}
\author[2]{Peng Li}

\affil[1]{\thepa{}{}}
\affil[2]{University of California, Santa Barbara}

\newcommand{\ourmethod}{\textsc{LASER}}

\begin{abstract}
  Accurate prediction of resource consumption and runtime for cloud workflow jobs is critical for scheduling efficiency, yet remains challenging due to the semi-structured nature of job configurations---comprising shell commands, tool-specific parameters, dependency graphs, and hierarchical metadata. Traditional ML approaches require brittle feature engineering to flatten this rich information into fixed-size vectors, losing critical semantic context. We present \ourmethod, a framework that fine-tunes LLMs on serialized workflow job configurations for multi-target resource and runtime regression. To address the challenges of numerical regression via generation, we introduce scientific notation output encoding for targets spanning multiple orders of magnitude, and constrained decoding with prefix filling to enforce output validity while reducing inference latency by over 30\%. We further show that full-attention fine-tuning improves accuracy over sliding-window LLMs on long job contexts. Validated on large-scale chip design workloads, and GHARuntime, a new public benchmark derived from 580{,}000+ GitHub Actions runs across 27{,}000+ repositories, \ourmethod\ outperforms human experts and SOTA tabular ML baselines, with clear model- and data-scaling behavior, establishing a new paradigm for LLM-based regression on semi-structured workflow data.
\end{abstract}

\begin{document}
\raggedbottom

\maketitle

\section{Introduction}

Accurate prediction of runtime and resource consumption for cloud workflow jobs is fundamental to efficient scheduling, cost control, and infrastructure planning. Mispredictions lead to either \emph{over-provisioning}---wasting expensive compute---or \emph{under-provisioning}---causing job failures and costly re-runs~\citep{liu2023cost, stok2014eda}. This challenge is particularly acute for modern workflow systems, which orchestrate complex pipelines of heterogeneous jobs across cloud and high-performance computing (HPC) infrastructures, where execution costs have skyrocketed as chip designs and scientific computations grow in complexity~\citep{bavikadi2022survey, ferreira2017future}. Workflows are typically structured as Directed Acyclic Graphs~(DAGs)~\citep{coleman2022wfcommons}, where each task---ranging from logic synthesis in large scale chip design simulations to sequence alignment in bioinformatics---carries a distinct configuration comprising tool invocations, command-line arguments, dependency specifications, and hardware constraints. Getting resource and runtime estimates right for such tasks is a prerequisite for backfill scheduling~\citep{liu2025ora}, makespan minimization~\citep{bader2024lotaru_journal}, and memory-efficient execution~\citep{lehmann2024ponder, bader2024sizey}.

Existing prediction approaches frame this as a tabular regression problem, requiring a feature-extraction function that maps complex job configurations into fixed-size vectors~\citep{liu2025ora, bader2024lotaru_journal}. Statistical learning methods use handcrafted features derived from numerical metadata such as submission time, requested runtime, and processor counts~\citep{feitelson2014pwa}, while deep learning methods extend this with temporal dependencies~\citep{liu2025ora}. However, this design forces practitioners to perform extensive and brittle feature engineering: command semantics (e.g., cross-compilation targets such as \texttt{aarch64} versus \texttt{x86\_64}), tool-specific parameters (e.g., \texttt{python-version: 3.11} versus \texttt{3.9}), and hierarchical structure (e.g., nested action configurations and inter-task dependencies) must all be hand-coded as scalar features~\citep{huang2021eda, zhu2024elastic}. Beyond being labor-intensive, such pipelines fail silently when new tools or commands appear---a common occurrence in rapidly evolving cloud services. Moreover, existing workflow-specific methods either rely on input-size-to-runtime linearity assumptions that do not hold universally in practice~\citep{lehmann2024ponder}, or require costly online profiling runs before prediction can begin~\citep{ bader2024lotaru_journal}. Critically, public workflow datasets such as WfCommons~\citep{coleman2022wfcommons} predominantly contain homogeneous task structures---e.g., the same binary applied to different input partitions---which further limits the scope of feature-based methods for predicting heterogeneous real-world workloads.

\begin{wrapfigure}{r}{0.67\textwidth}
  \vspace{-0.8em}
  \centering
  \includegraphics[width=0.67\textwidth,trim=6cm 7.3cm 14cm 0.5cm,clip]{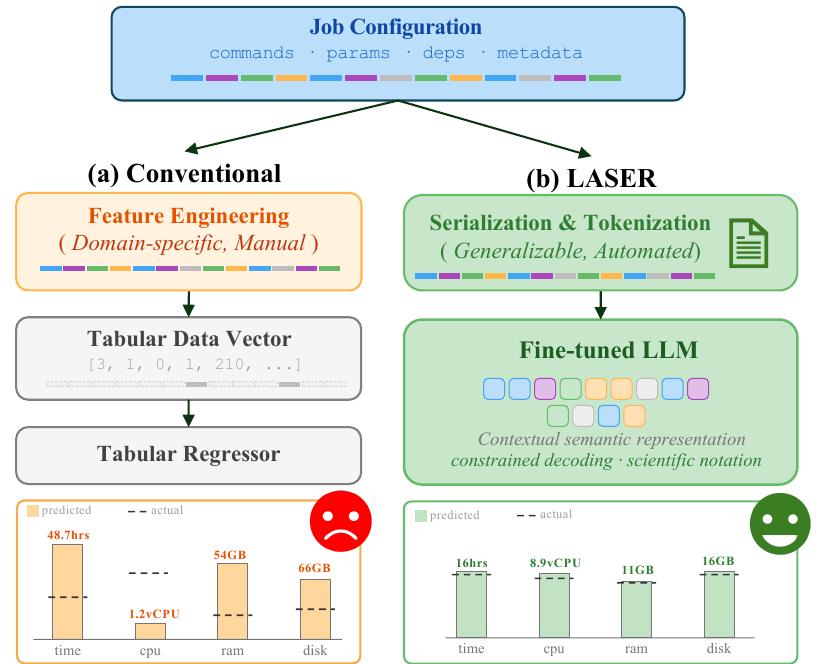}
  \vspace{-0.8em}
  \caption{Overview of conventional tabular prediction vs. LASER for workflow job resource and runtime estimation.}
  \label{fig:motivation}
  % \vspace{-0.8em}
\end{wrapfigure}

Large Language Models~(LLMs) offer a fundamentally different approach. Having been pre-trained on vast corpora that include shell scripts, configurations, build systems, and domain-specific toolchains, LLMs can encode semantic signals from raw, semi-structured job configurations without any manual feature engineering. Recent work has demonstrated that LLMs can serve as effective regressors when fine-tuned appropriately through text-to-number generation~\citep{song2024omnipred, song2025decoding, akhauri2025regression}. Concurrently, \citet{liu2025ora} showed that LLMs enhanced with retrieval-augmented generation~(RAG) can leverage job script content to improve HPC runtime prediction---yet their approach requires per-query retrieval from a historical database, incurring significant inference latency, and cannot generalize to jobs unseen by the database. This opportunity for direct fine-tuning of LLMs on semi-structured workflow job configurations remains entirely unexplored.

We present \ourmethod, a framework for \textbf{L}anguage model-b\textbf{A}sed regre\textbf{S}sion for s\textbf{E}mi-structured workflow job \textbf{R}esource and runtime estimation. \ourmethod\ fine-tunes LLMs on serialized workflow job configurations for simultaneous prediction of wall-clock runtime, peak CPU, peak memory, and peak disk usage, treating the problem as a sequence-to-sequence generation task. Three techniques make this effective. First, we represent all numerical targets in scientific notation, providing a normalization-free, compact encoding for values spanning multiple orders of magnitude (e.g., memory from megabytes to terabytes) that avoids the training instability of linear regression heads~\citep{song2024omnipred}. Second, we introduce constrained decoding with deterministic prefix filling, which enforces structural output validity, eliminates hallucinations such as malformed JSON or fabricated keys, and reduces inference latency by over 30\% by bypassing forward passes for deterministic structural tokens. Third, we demonstrate that full-attention fine-tuning substantially outperforms the default sliding-window attention of modern LLMs on long job configurations, where global context across distant command arguments and dependency specifications is critical for accurate prediction.

We validate \ourmethod\ on two types of workflows. The first is a proprietary industrial chip design verification (ChipDV) workflow, comprising over 160{,}000 cloud jobs with rich semi-structured configurations of design verification workloads---domains where a minor change in a synthesis script can cause a tenfold increase in the runtime of the subsequent stage~\citep{huang2021eda, zhu2024elastic}. The second is GHARuntime, a new public benchmark we derive from the GHALogs corpus~\citep{moriconi2025ghalogs}, containing over 1.3 million continuous integration and continuous deployment (CI/CD) workflow job execution records across 27{,}000+ GitHub repositories spanning 20 programming languages. Detailed backgrounds about these two datasets can be found in Appendix~\ref{sec:background}.
% Unlike prior workflow trace collections such as WfCommons~\citep{coleman2022wfcommons}, which aggregate scientific workflows with homogeneous task structures, GHARuntime captures \emph{genuinely heterogeneous} job configurations: each job comprises sequences of GitHub Actions with tool-specific parameters and shell commands that vary widely across repositories and programming ecosystems, with job durations spanning three orders of magnitude. On both datasets, fine-tuned \ourmethod\ models significantly outperform production heuristic baselines and tabular ML methods, with prediction accuracy scaling predictably with model size from 270M to 12B parameters.

In summary, we make the following contributions:
\begin{itemize}[leftmargin=*]

\item We present the first framework for fine-tuning LLMs directly on semi-structured workflow job configurations for runtime and multi-resource prediction, eliminating the need for brittle feature engineering and outperforming both production heuristics and tabular ML baselines.

\item We introduce scientific notation output encoding and constrained decoding with deterministic prefix filling, jointly improving prediction accuracy, output reliability, and inference efficiency by over 30\%.

\item We release \textbf{GHARuntime}, a large-scale public benchmark derived from the GHALogs dataset~\citep{moriconi2025ghalogs}, comprising 1.3M+ CI/CD workflow job records with rich command-level metadata across 27K repositories, enabling reproducible research on workflow job runtime prediction.

% \item We demonstrate consistent improvements over baselines across two datasets---one industrial, one public--- with clear LLM scaling behavior across model sizes, and temporal out-of-distribution generalization over a 26-day horizon.

\end{itemize}

\begin{figure*}[t]
  \centering
  \includegraphics[width=\linewidth]{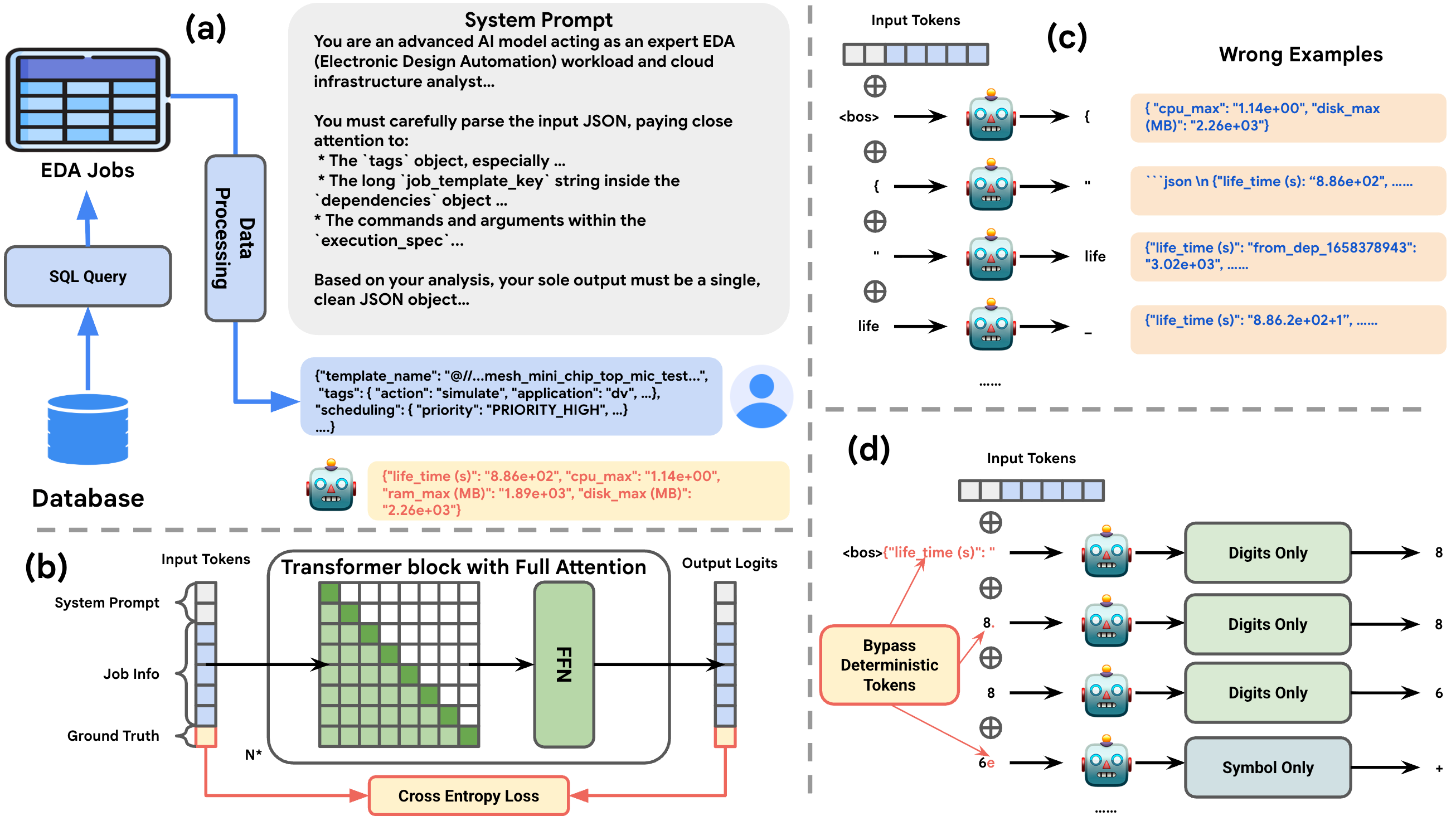}
  \caption{Architecture of \ourmethod\ for chip design verification workload prediction: (a) The data flow and conversational example. (b) Supervised fine-tuning LLM with full attention mechanism to minimize CE loss on ground truth tokens. (c) Vanilla LLM decoding methods and examples of wrong generations. (d) The proposed constrained decoding method.}
  \label{fig:system-diagram}
\end{figure*}

% \section{Preliminaries}

% \input{sections/preliminaries}

\section{Related Work}
\paragraph{Language Model Regression.}
Treating regression as sequence generation has recently emerged as a viable alternative to classical regression heads. LLMs fine-tuned on $(x,y)$ text pairs can outperform traditional regressors across diverse domains in Google Vizier's blackbox optimization database~\citep{song2024omnipred}. Theoretical work shows that cross-entropy-trained decoders over tokenized numbers match pointwise regression heads in expectation~\citep{song2025decoding}. LLMs can also perform non-linear regression via in-context learning without parameter updates, rivaling Random Forest and Gradient Boosting~\citep{vacareanu2024words}. Most relevant to our setting, text-to-text regression was applied to Google's Borg cluster, achieving 100$\times$ lower MSE than tabular baselines on resource efficiency prediction~\citep{akhauri2025large}, and was later extended to code-to-metric regression across 17 languages and GPU kernels using a unified Regression Language Model~\citep{akhauri2025regression}, with a encoder-decoder architecture. \textsc{LASER} differs from this line of work by algining \emph{decoder-only} LLMs on semi-structured workflow JSON with constrained numeric output generation and multi-target prediction.

\paragraph{Workflow and HPC Job Runtime Prediction.}
Classical HPC runtime prediction relies on numerical metadata from batch traces~\citep{feitelson2014pwa}, with LightGBM emerging as the strongest tabular baseline~\citep{chen2022lightgbm}. ORA augments LLMs with retrieval over job scripts and metadata, improving accuracy by over 40\% versus metadata-only ML baselines~\citep{liu2025ora}. Unlike ORA, \textsc{LASER} uses supervised fine-tuning for single-pass inference without retrieval, generalizing to cold-start jobs from unseen repositories. For scientific workflow tasks, runtimes have been predicted via local profiling and Bayesian linear regression, assuming linear input-size-to-runtime relationships that rarely hold for EDA or CI/CD jobs~\citep{bader2024lotaru_journal}. Online memory sizing for Nextflow workflows has also been addressed by selecting among multiple ML models at the task-type level~\citep{bader2024sizey,lehmann2024ponder}---a setting of repeated homogeneous tasks that is distinct from the heterogeneous job configurations in our datasets.

\paragraph{Workflow Datasets.}
WfCommons~\citep{coleman2022wfcommons} provides scientific workflow traces in standardized JSON, but most workflows consist of the same binary applied to different input partitions, offering limited semantic diversity for text-based methods. The Parallel Workload Archive~\citep{feitelson2014pwa} contains HPC batch traces in purely numerical SWF format with no script information. Our \textbf{GHARuntime} benchmark, derived from GHALogs~\citep{moriconi2025ghalogs}, addresses both gaps with 1.3M CI/CD job records across 27K repositories featuring rich command-level metadata---to our knowledge the first public benchmark for workflow runtime prediction at this scale and semantic richness.

\section{Methodology}

\begin{figure*}[t]
    \centering
    \begin{subfigure}[b]{0.48\textwidth}
         \centering
         \includegraphics[width=\textwidth]{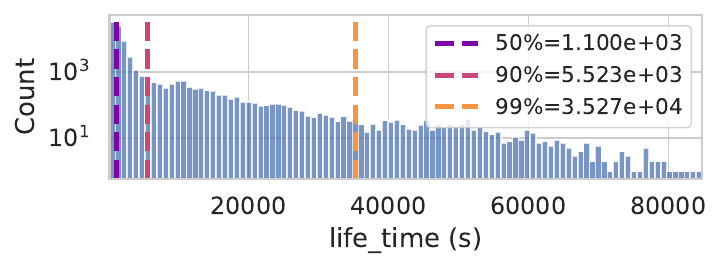}
    \end{subfigure}
    \begin{subfigure}[b]{0.48\textwidth}
         \centering
         \includegraphics[width=\textwidth]{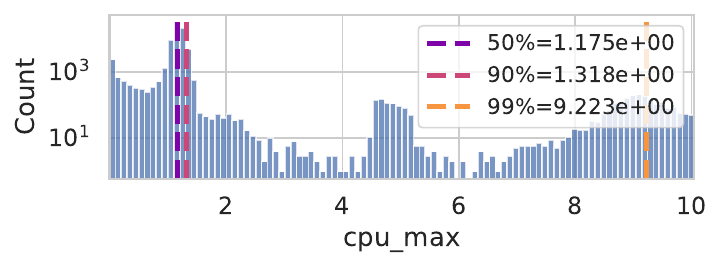}
    \end{subfigure}
    \hfill
    \begin{subfigure}[b]{0.48\textwidth}
         \centering
         \includegraphics[width=\textwidth]{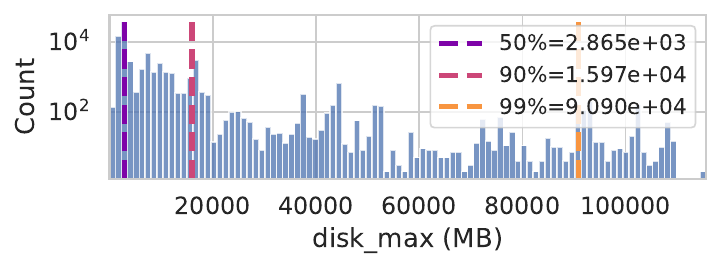}
    \end{subfigure}
    \begin{subfigure}[b]{0.48\textwidth}
         \centering
         \includegraphics[width=\textwidth]{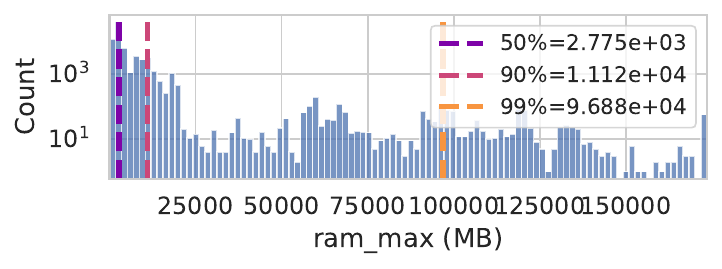}
    \end{subfigure}
    \caption{Histograms illustrating the distribution of key resource metrics from dataset 1. All distributions are heavily right-skewed, indicating a long tail of resource-intensive jobs.}
    \label{fig:hist-metric}
\end{figure*}

We frame resource and runtime prediction as a sequence-to-sequence generation task: given a serialized job configuration as input, the model generates a structured output containing predictions for wall-clock runtime, peak CPU, peak memory, and peak disk usage. This formulation lets us leverage the full representational capacity of pre-trained LLMs and the training stability of Cross-Entropy loss $\mathcal{L}_{\text{CE}}$ to avoid the instability of linear regression heads over targets spanning multiple orders of magnitude. The key components of \ourmethod: input serialization, numerical encoding,  fine-tuning, and constrained decoding. These components are illustrated in \cref{fig:system-diagram} and detailed in the following sections.

\paragraph{Input and Output Representation}
Let $\mathbf{x} \in \mathcal{X}$ denote a raw job configuration and $\mathbf{y} = [y^{(1)}, \dots, y^{(K)}] \in \mathbb{R}_{>0}^{K}$ the corresponding resource targets. We define a serialization function $\phi : \mathcal{X} \to \mathcal{V}^{*}$ that converts each configuration into a unified JSON token sequence, ordering fields by descending predictive importance so that the most informative content survives right-truncation to $L_{\max}$ tokens (ablated in \cref{subsec:max-seq-len}). Targets are numerically encoded (\cref{subsec:sci-notation}) and serialized as a structured JSON output. We set $L_{\max}{=}2048$ for all experiments.

\subsection{Scientific Notation Encoding}\label{subsec:sci-notation}

Resource metrics exhibit heavy right-skewed distributions spanning multiple orders of magnitude. Take our ChipDV dataset for an instance, \texttt{ram\_max} ranges from the 50th percentile of $2.8\times10^3$ MB to the 99th percentile of $9.7\times10^4$ MB, while \texttt{life\_time} spans from $1.1\times10^3$s to $3.5\times10^4$s in \cref{fig:hist-metric}. This long-tailed variation poses a fundamental challenge for regression heads: a linear or LogNormed MLP head fails to generalize across such ranges, and while output normalization partially mitigates this, it requires per-target, per-domain scaling that constitutes its own form of brittle feature engineering and balancing multi-objective optimization \citep{akhauri2025regression}. 

We leverage scientific notation to transform the regression problem into a next-token prediction problem. We encode each scalar target $y^{(k)} \in \mathbb{R}_{>0}$ as:
\begin{equation}
    y^{(k)} = m^{(k)} \times 10^{\,e^{(k)}}, \qquad m^{(k)} \in [1.00,\; 9.99], \quad e^{(k)} \in \mathbb{Z},
\end{equation}
$y^{(k)}$ is a fixed-length sequence of $L_{\text{fix}}{=}8$ tokens from $\mathcal{V}_{\text{num}} = \{\texttt{0},\dots,\texttt{9},\texttt{.},\texttt{e},\texttt{+},\texttt{-}\}$. The full prediction target $\mathbf{y}$ is then serialized as a JSON object with all $K$ metrics encoded in this form, e.g., \texttt{\{"life\_time (s)": "8.86e+02", "cpu\_max": "1.14e+00", ...\}}. This gives two benefits: the mantissa is always bounded in $[1,10)$, making per-target normalization unnecessary; and the fixed output length per metric simplifies constrained decoding (\cref{subsec:constrain-decoding}) and reduces generation cost.

\textit{Remark.}
Some LLM tokenizers split floating-point literals into variable-length, semantically opaque subword tokens, making it difficult to learn consistent numerical representations across orders of magnitude---a critical issue when targets range from megabytes to terabytes. We find that Qwen-3 \citep{qwen-3} and Gemma-3 \citep{gemma3} tokenizers encode each digit \texttt{0}--\texttt{9} as a single token, a property required for fixed-length scientific notation encoding. LLaMA-3, by contrast, merges digit sequences into multi-character tokens, violating this assumption. We therefore restrict our experiments to Qwen-3 and Gemma-3 architectures.

\subsection{Supervised Fine-Tuning with Full Attention}\label{subsec:sft}

Given a dataset of $N$ job configurations and their corresponding resource targets $\mathcal{D} = \{(\mathbf{x}_i, \mathbf{y}_i)\}_{i=1}^{N}$, we fine-tune a pre-trained LLM to generate the structured JSON output autoregressively. Each training example is formatted as a prompt-completion pair: the prompt contains the serialized job configuration $\tilde{\phi}(\mathbf{x}_i)$ wrapped in a standard instruction template, and the completion is the JSON-serialized target $\mathbf{y}_i$ with all numerical values encoded in scientific notation (\cref{subsec:sci-notation}). We minimize the cross entropy (CE) loss over completion tokens only:
\begin{equation}
    \mathcal{L}_{\text{CE}} = -\frac{1}{|\mathcal{T}|}\sum_{t \in \mathcal{T}} \log p_\theta\left(w_t \mid w_{<t}\right),
\end{equation}
where $\mathcal{T}$ indexes the completion tokens and $p_\theta$ is the LLM parameterized by $\theta$.

\paragraph{Full-Attention Fine-Tuning.} Modern long-context LLMs default to sliding-window attention to reduce computational cost, restricting each token's receptive field to a local window. However, workflow job configurations not only have long contexts, varying hundrads to 70k tokens of ChipDV workloads, but also encode predictive signals across distant fields, a compilation target specified early in the configuration may critically determine the runtime of a tool invoked much later. We find that replacing sliding-window attention with full attention during fine-tuning substantially improves prediction accuracy on long job contexts, and ablate this choice in \cref{subsec:ablation}.

% We implement the proposed formulation using a decoder-only Transformer architecture as  in \cref{fig:system-diagram}(b). The training process involves constructing a comprehensive context sequence $\mathbf{S}$ by concatenating three segments: (1) system prompt ($\mathbf{p}_{\text{sys}}$), instructing the model on the role and output format; (2) serialized job information ($\mathbf{x}_{\text{job}}$), derived from the raw configuration $\mathbf{x}$; (3) ground truth metrics ($\mathbf{y}_{\text{gt}}$), serving as the training target.

% Let the input context be denoted by $\mathbf{C} = [\mathbf{p}_{\text{sys}}; \mathbf{x}_{\text{job}}]$. We apply Full Attention across the entire sequence length $N^*$. The model is optimized end-to-end by minimizing the autoregressive cross-entropy loss $\mathcal{L}_{\text{CE}}$ specifically on the tokens corresponding to the resource metrics:
% \begin{equation}
%     \mathcal{L}_{\text{CE}} = - \sum_{t=1}^{T} \log P(y_t \mid \mathbf{C}, y_{<t})
% \end{equation}
% where $y_t$ represents the $t$-th token of the ground truth metrics $\mathbf{y}_{\text{gt}}$, and $y_{<t}$  the preceding metric tokens. This masking strategy ensures the model learns the causal correlation between workload parameters defined in $\mathbf{C}$ and the resulting resource consumption.

% Requires \usepackage[table]{xcolor} (loads \texttt{colortbl}) for \rowcolor.
% If colors are already defined in the preamble, remove the next two lines.
\definecolor{lightblue}{HTML}{E3F2FD}     % light blue
\definecolor{lightgreen}{HTML}{E8F5E9}    % light green
\begin{table*}[!t]
    \centering
\small
    % Global setup for the MAE columns
    \sisetup{scientific-notation=false, round-mode=places, round-precision=2}
    \resizebox{\textwidth}{!}{ % Forces table to fit the column width

    \begin{tabular}{
        l % Method
        c % Life (min)
        c % CPU
        c % RAM (GB)
        c % Disk (GB)
        c % Pearson Avg
        c % Spearman Avg
        }
    \toprule
    \multirow{3}{*}{\textbf{Method}} & \multicolumn{4}{c}{\textbf{Test MAE ($\downarrow$)}} & \multicolumn{1}{c}{\textbf{Pearson}} & \multicolumn{1}{c}{\textbf{Spearman}} \\
    \cmidrule(lr){2-5}
     & \multicolumn{1}{c}{\textbf{life\_time}} & \multicolumn{1}{c}{\textbf{cpu\_max}} & \multicolumn{1}{c}{\textbf{ram\_max}} & \multicolumn{1}{c}{\textbf{disk\_max}} & \multicolumn{1}{c}{\textbf{Avg.}} & \multicolumn{1}{c}{\textbf{Avg.}} \\
     & \multicolumn{1}{c}{\textbf{(min)}} & \multicolumn{1}{c}{\textbf{(vCPU)}} & \multicolumn{1}{c}{\textbf{(GB)}} & \multicolumn{1}{c}{\textbf{(GB)}} & \multicolumn{1}{c}{\textbf{($\uparrow$)}} & \multicolumn{1}{c}{\textbf{($\uparrow$)}} \\
    \midrule
    \rowcolor{lightblue}\multicolumn{7}{c}{\textit{ChipDV-Repetitive}} \\
    \rowcolor{lightblue}& & & & & & \\[-7pt]
   \rowcolor{lightblue}Human Expert          & {---}   & 0.311 & 54.90 & 66.70 & {---} & {---} \\
    \rowcolor{lightblue}Historical Max             & {---}   & 0.295 & 16.20 & 21.70 & {---} & {---} \\
    % \midrule
    \rowcolor{lightblue}& & & & & & \\[-7pt]
   \rowcolor{lightblue}\multicolumn{7}{l}{\ourmethod:} \\
    \rowcolor{lightblue}\quad Gemma-3-270M          & 69.29  & 0.301 & 11.90 & 7.14  & 0.469 & 0.509 \\
    \rowcolor{lightblue}\quad Gemma-3-1B            & 51.30  & 0.183 & 3.29  & 1.39  & 0.834 & 0.773 \\
    \rowcolor{lightblue}\quad Gemma-3-4B            & 47.02  & 0.168 & 2.61  & 1.08  & 0.829 & 0.821 \\
    \rowcolor{lightblue}\quad Gemma-3-12B           & \textbf{38.02}  & \textbf{0.128} & 1.89  & 0.80  & 0.913 & \textbf{0.869} \\
    \rowcolor{lightblue}\quad Qwen-3-8B             & 41.07  & 0.130 & \textbf{1.66}  & \textbf{0.76}  & \textbf{0.919} & 0.858 \\
    \midrule
    \rowcolor{lightgreen}\multicolumn{7}{c}{\textit{ChipDV-Unseen}} \\
    \rowcolor{lightgreen}& & & & & & \\[-7pt]
   \rowcolor{lightgreen}Human Expert    & {---}   & 0.330 & 53.60 & 59.08 & {---}  & {---} \\
    % \midrule
    \rowcolor{lightgreen}& & & & & & \\[-7pt]
   \rowcolor{lightgreen}\multicolumn{7}{l}{\ourmethod:} \\
    \rowcolor{lightgreen}\quad Gemma-3-270M    & 37.73  & 0.478 & 3.45  & 3.75  & 0.502 & 0.438 \\
    \rowcolor{lightgreen}\quad Gemma-3-1B      & 27.91  & 0.165 & 2.30  & 1.30  & 0.767 & 0.701 \\
    \rowcolor{lightgreen}\quad Gemma-3-4B      & 26.25  & 0.169 & 1.56  & 0.95  & 0.832 & 0.745 \\
    \rowcolor{lightgreen}\quad Gemma-3-12B     & \textbf{23.76}  & 0.145 & \textbf{1.40}  & 1.14  & \textbf{0.858} & 0.799 \\
    \rowcolor{lightgreen}\quad Qwen-3-8B       & 24.52  & \textbf{0.136} & 1.41  & \textbf{0.94}  & 0.856 & \textbf{0.804} \\
    \bottomrule
    \end{tabular}
    }
    \caption{Test results on industrial ChipDV workflow. {---} indicates the baseline does not produce a prediction for that target.}
    \label{tab:results_combined}
\end{table*}

\begin{figure*}[!t]
  \centering
  \includegraphics[width=0.98\linewidth]{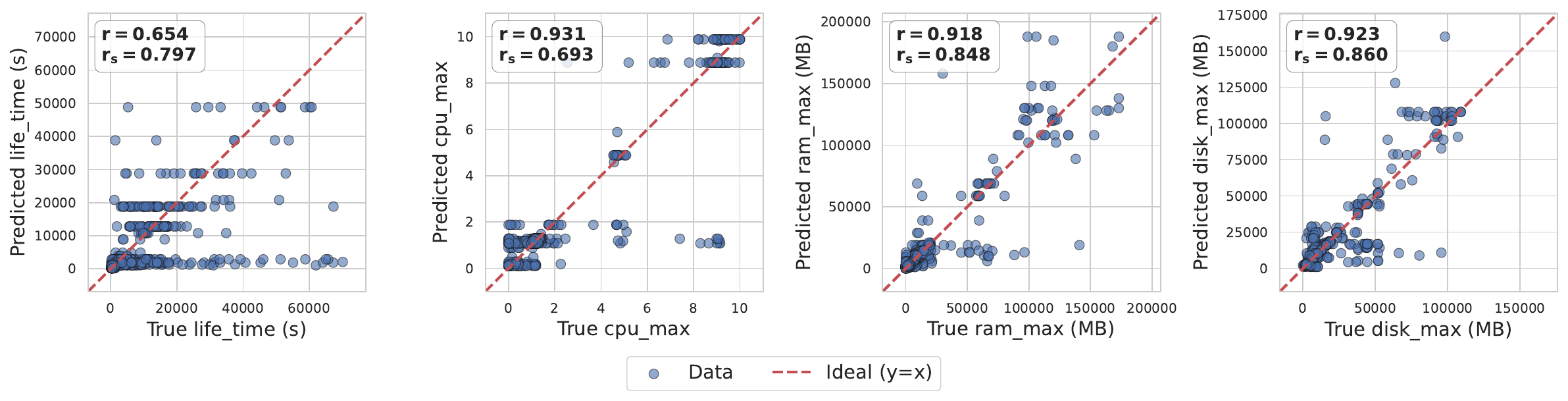}
  \caption{True vs. predicted resource values from the Gemma-3-12B model on the test samples of dataset 1.
    % The red dashed line represents the ideal (y=x) case where the predicted value equals the true value. The scatter points' proximity to this line, along with the high Pearson ($r$) and Spearman ($r_s$) correlation coefficients shown in the upper-left of each plot, demonstrates the model's high predictive accuracy.
  }    \label{fig:scatter-gemma-12b-dst-1}
\end{figure*}

\subsection{Constrained Decoding with Prefix Filling} \label{subsec:constrain-decoding}

At inference time, the model must generate a well-formed JSON object whose values are valid scientific notation strings. Vanilla autoregressive decoding provides no such guarantee: the model may hallucinate malformed keys, emit invalid numerical formats, or produce structurally inconsistent outputs (see \cref{fig:system-diagram}(c)). We address this with constrained decoding with deterministic prefix filling.

The key observation is that the output structure is fully determined at inference time: the JSON keys, delimiters, and scientific notation scaffolding (e.g., \texttt{\{"life\_time (s)": "<mantissa>e<exp>", ...\}}) are known constants, leaving only the mantissa digits and exponent as free variables. We therefore partition the output token sequence into \emph{deterministic} tokens---structural tokens whose values are fixed by the output schema---and \emph{generative} tokens---the $K \times L_{\text{fix}}$ numerical digits produced by the model. Deterministic tokens are filled directly without invoking a forward pass, while generative tokens are constrained to $\mathcal{V}_{\text{num}}$ via logit masking.

This yields two complementary benefits. First, output validity is enforced by construction: malformed JSON and invalid numerical formats are structurally impossible. Second, skipping forward passes for all deterministic tokens reduces inference latency by over 30\%, a saving that grows with the number of predicted metrics $K$.

Finally, we observe that LLM may generate extreme values that are far beyond any training examples, due to our-of-range exponent digits. So we further clip model's output to fit within the target value range of training examples. 

% On large-scale EDA cloud infrastructure, the volume of workload processing is immense, often reaching millions of jobs daily. Consequently, even a trivial failure rate in result generation can cascade into significant operational disruptions or data pipeline failures. Standard autoregressive decoding, as illustrated in \cref{fig:system-diagram}(c), lacks structural guarantees; it is prone to "hallucinations" such as malformed JSON syntax, the fabrication of non-existent keys, or the generation of irrelevant conversational text. To eliminate these stochastic instabilities and ensure strictly parseable outputs, we implement a \textit{Constrained Decoding} framework that enforces structural rigidity. As depicted in \cref{fig:system-diagram}(d), our decoding strategy partitions the generation process into two distinct phases: \textit{Deterministic Prefix Bypass} and \textit{Constrained Numeric Sampling}.

% \subsubsection{Deterministic Prefix Bypass}
% To maintain schema consistency, we treat the JSON keys and structural delimiters as deterministic templates. Instead of allowing the model to predict these static tokens, we employ a \textit{generation prefix filling} technique. During inference, deterministic prefixes such as \verb|{"life_time (s)": "8.86e+02"}| are automatically appended to the input context window, as shown in (d) of \cref{fig:system-diagram}. 

\begin{table*}[!t]
\centering
\sisetup{scientific-notation=false, round-mode=places, round-precision=2}
\resizebox{\textwidth}{!}{%
\begin{tabular}{
    l
    c c c
    c c c
    c c c
    }
\toprule
  & \multicolumn{9}{c}{\textbf{Test life\_time (second)}} \\
\cmidrule(lr){2-10}
\textbf{Dataset} 
 & \multicolumn{3}{c}{\textbf{GHARuntime-1k}}
 & \multicolumn{3}{c}{\textbf{GHARuntime-10k}}
 & \multicolumn{3}{c}{\textbf{GHARuntime-100k}} \\
\cmidrule(lr){2-4} \cmidrule(lr){5-7} \cmidrule(lr){8-10}
\multirow{2}{*}{\textbf{Method}}
 & \textbf{MAE} & \textbf{Pearson} & \textbf{Spearman}
 & \textbf{MAE} & \textbf{Pearson} & \textbf{Spearman}
 & \textbf{MAE} & \textbf{Pearson} & \textbf{Spearman} \\
 & \textbf{($\downarrow$)} & \textbf{($\uparrow$)} & \textbf{($\uparrow$)}
 & \textbf{($\downarrow$)} & \textbf{($\uparrow$)} & \textbf{($\uparrow$)}
 & \textbf{($\downarrow$)} & \textbf{($\uparrow$)} & \textbf{($\uparrow$)} \\
\midrule
XGBoost %\citep{chen2016xgboost}
  & 645.20 & 0.02 & 0.09
  & 485.03 & 0.21 & 0.58
  & 487.01 & 0.21 & 0.60 \\
LightGBM %\citep{ke2017lightgbm}
  & 812.13 & $-0.10$ & $-0.06$
  & 464.73 & 0.23 & 0.56
  & 468.43 & 0.28 & 0.52 \\
TabPFN-v2.6 %\citep{hollmann2025tabpfn}
  & \textbf{333.53} & \textbf{0.29} & \textbf{0.55}
  & 325.88 & \textbf{0.28} & \textbf{0.67}
  & 340.18 & \textbf{0.33} & 0.67 \\
\addlinespace[3pt]

\multicolumn{10}{l}{\ourmethod:} \\
\quad Qwen-3-0.6B
  & 7324.92 & $-0.08$ & 0.25
  & 365.35 & 0.12 & 0.44
  & 326.30 & 0.28 & 0.67 \\
\quad Qwen-3-8B
  & 4001.35 & 0.27 & 0.50
  & \textbf{308.25} & 0.24 & 0.60
  & \textbf{312.20} & \textbf{0.33} & \textbf{0.72} \\
\bottomrule
\end{tabular}%
}
\caption{GHARuntime test metrics of job lifetime.}
\label{tab:gharuntime_results}
\end{table*}

% \subsubsection{Constrained Numeric Sampling}
% Once the deterministic prefix is supplied, the model must generate the specific numerical values. To prevent the generation of non-numerical text, we dynamically restrict the model's vocabulary space $\mathcal{V}$.

% In the logical flow of \cref{fig:system-diagram}(d), we apply a masking function $\mathcal{M}$ to the output logits at each step $t$ where a numerical value is expected. We define a valid token subset $\mathcal{V}_{num} \subset \mathcal{V}$, which consists strictly of digit tokens form \texttt{<0>} to \texttt{<9>}. During the generation of value fields, the probability of any token $w_i \notin \mathcal{V}_{num}$ is set to zero. Similarly, we can sample symbol tokens from \texttt{<+>} and \texttt{<->}.
% % \begin{equation}
% %     P(w_i | x) = 
% %     \begin{cases} 
% %       \frac{\exp(z_i)}{\sum_{j \in \mathcal{V}_{num}} \exp(z_j)} & \text{if } w_i \in \mathcal{V}_{num} \\
% %       0 & \text{otherwise}
% %     \end{cases}
% % \end{equation}
% This  ensures that the output is always a valid number in scientific notation, effectively reducing the parsing error rate to zero.

% Finally, we observe that LLM may generate extreme values that are far beyond any training examples. So we further clip model's output to fit within the target value range of training examples. 

\section{Experimental Results}

% We summarize the setup here and defer complete baseline definitions,
% dataset statistics, and training hyperparameters to
% Appendix~\ref{app:exp-setup}. We then present the main results and
% ablation studies.

% \subsection{Experimental Setup}
% \label{sec:exp-setup}

Experiments are conducted on (1) proprietary industrial ChipDV workflow, including multi-targerts of job life time, peak usage of virtual CPU (vCPU)\footnote{In our cloud platform, a vCPU is a virtualization abstraction representing a single hardware multithread on an underlying physical CPU processor.}, memory, and disk; and (2) our derived GHARuntime workflow, a single target prediction task of job runtime. Due to industrial privacy constraints, ChipDV configuration details cannot be disclosed. \ourmethod's system prompt and GHARuntime examples are provided in Appendix~\ref{app:exp-setup}.

\begin{wraptable}[9]{r}{0.58\columnwidth}
\centering
% \scriptsize
\setlength{\tabcolsep}{4pt}
\begin{tabular}{l c c c}
\toprule
\textbf{Gemma-3} & \textbf{1B} & \textbf{4B} & \textbf{12B} \\
\midrule
Vanilla     & 2.03 & 3.42 & 5.71 \\
Constrained & 1.38 & 2.18 & 3.99 \\
Reduction   & $-$32.4\% & $-$36.1\% & $-$30.1\% \\
\bottomrule
\end{tabular}
\caption{Inference time (s) for vanilla vs.\ constrained decoding.}
\label{tab:constrained_decoding}
\end{wraptable}
We fine-tune Gemma-3 and Qwen-3 models with LoRA \citep{lora}: we set LoRA rank as 8, LoRA alpha as 16, and use a dropout rate of 0.05. We train for 2 epochs with maximum sequence length 2048, learning rate $2\times10^{-5}$, and batch size 64. During testing generation, we set \ourmethod\ LLM's temperature to 0, leading greedy decoding to eliminate response randomness. We report mean absolute error (MAE), Pearson correlation ($r$), and Spearman correlation ($r_s$).
% Detailed setup choices, including data collection, sampling protocol, and exact optimization hyperparameters, are provided in Appendix~\ref{app:exp-setup}.

We evaluate \ourmethod\ against human experts, historical maximum records on ChipDV workloads; for GHARuntime, we compare our approach with competitive tabular regressors: XGBoost \citep{chen2016xgboost}, Light GBM \citep{ke2017lightgbm}, and TabPFN-v2.6 \citep{hollmann2025tabpfn} \footnote{TabPFN-v2.6 checkpoint is released on March 24th 2026 along with no technical report or paper. We cite the journal version of TabPFN to compromise this situation.}.

% \subsection{Main Results}

\subsection{In-Distribution Performance}
We evaluate \ourmethod\ on two datasets sampled from the ChipDV workflow database with different schema: ChipDV-Repetitive (87{,}616 jobs, $\approx$64\% ones matching historical records), and a more challenging one ChipDV-Unseen (74{,}968 jobs, none of which can be found in the database). both dataset jobs are uniformly sampled from the same chip project within 10-day time window. We further slpit 80\% data for training, 10\% for validation, and 3{,}000 held-out examples for benckmarking, leading training and testing data to follow the same distribution.

As shown in \cref{tab:results_combined}, \ourmethod\ significantly outperform human experts and historical check across both datasets. \cref{fig:scatter-gemma-12b-dst-1} illustrates Gemma-3-12B predictions on ChipDV-Unseen, demonstrating \ourmethod's generalization across differnt metric value scales. We further observe the scaling law between model size and prediction accuracy: the two largest models, Gemma-3-12B and Qwen-3-8B, achieve the best performance across all targets.

\subsection{Out of Distribution (OOD) Performance}

A critical requirement for workflow resource predictors is generalization beyond the training distribution—either to future jobs under temporal drift, or to novel workflows from unseen codebases. We evaluate \ourmethod\ on two OOD tasks:
\begin{itemize}[leftmargin=*]
  \item \textbf{Temporal drift}: fine-tuned on ChipDV-Repetitive, tested on future ChipDV jobs partitioned into 5-day windows (up to 26 days ahead);
  \item \textbf{Cross-repository generalization}: fine-tuned on a subset of GHARuntime repositories, tested on held-out repositories.
\end{itemize}

\textbf{Temporal Drift.}
Table~\ref{tab:temporal-drift} reports MAE of Gemma-3-4B across six 5-day future windows. Performance on \texttt{cpu\_max}, \texttt{ram\_max}, and \texttt{disk\_max} remains stable throughout the 26-day period, demonstrating strong resistance to temporal drift. \texttt{life\_time} MAE is higher in the 6--10 day window but recovers in later windows, suggesting the model has learned generalizable resource consumption patterns rather than overfitting to the training period distribution.

\textbf{Cross-Repository Generalization.}
Table~\ref{tab:gharuntime_results} reports results on GHARuntime across three dataset scales (1k, 10k, 100k training jobs and 20\% testing jobs). Two complementary scaling trends emerge. Beyound model-scaling improvement from Qwen-3-0.6B to Qwen-3-8B, \ourmethod\ scales with data size as well: at 1k training samples both variants underperform ML baselines, but at 10k--100k they match or surpass TabPFN, v2.6—the strongest baseline. This data-scaling behavior highlights that \ourmethod\ requires sufficient fine-tuning signal to adapt to workflow resource estimation task, while ML methods plateau early and fail to improve further with more data.

\begin{table*}[!t]
  \centering
  \begin{minipage}[t]{0.47\textwidth}
    \vspace{0pt}
    
\centering
\scriptsize
\resizebox{\columnwidth}{!}{ % Uncomment to force table to fit column width
\begin{tabular}{l c c c c}
\toprule
\multirow{3}{*}{\textbf{Day Shift}} & \multicolumn{4}{c}{\textbf{Test MAE ($\downarrow$)}} \\
\cmidrule(lr){2-5}
 & \multicolumn{1}{c}{\textbf{life\_time}} & \multicolumn{1}{c}{\textbf{cpu\_max}} & \multicolumn{1}{c}{\textbf{ram\_max}} & \multicolumn{1}{c}{\textbf{disk\_max}} \\
 & \multicolumn{1}{c}{\textbf{(min)}} & \multicolumn{1}{c}{\textbf{(vCPU)}} & \multicolumn{1}{c}{\textbf{(GB)}} & \multicolumn{1}{c}{\textbf{(GB)}} \\
\midrule
0     & 47.02 & 0.168 & 2.61 & 1.08  \\
1--5  & 62.19 & 0.154 & 3.06 & 1.79 \\
6--10 & 76.86 & 0.157 & 3.37 & 1.73 \\
11--15& 30.27 & 0.161 & 2.92 & 1.52 \\
16--21& 33.36 & 0.173 & 2.71 & 1.52 \\
22--26& 44.65 & 0.146 & 2.50 & 1.78 \\
\bottomrule
\end{tabular}
} % Uncomment this to close \resizebox
\caption{Temporal OOD performance of \ourmethod\ (Gemma-3-4B) on ChipDV, evaluated on non-overlapping future 5-day windows.}
\label{tab:temporal-drift}

  % \end{minipage}\hfill
  % \begin{minipage}[t]{0.2\textwidth}
  %   \vspace{0pt}
  %   \input{tables/constrained_decoding}
  \end{minipage}\hfill
  \begin{minipage}[t]{0.47\textwidth}
    \vspace{0pt}
    % \begin{table*}[h!]
% \centering
% \caption{Correlation between Test CE Loss and Test MAE during training on Dataset 2. A consistent decrease in CE Loss corresponds to a decrease in MAE across all metrics.}
% \label{tab:ce_mae_correlation}
% \sisetup{scientific-notation=true, round-mode=places, round-precision=2}
% \begin{tabular}{cccccc}
% \toprule
% \textbf{Epoch} & \textbf{Validation CE Loss ($\downarrow$)} & \multicolumn{4}{c}{\textbf{Test MAE ($\downarrow$)}} \\
% \cmidrule(lr){3-6}
%  & & \textbf{life\_time (s)} & \textbf{cpu\_max (vcpu)} & \textbf{ram\_max (GB)} & \textbf{disk\_max (GB)} \\
% \midrule
% 0.5 & 0.2873 & \num{3.66e+03} & \num{.253} & \num{6.19} & \num{3.24} \\
% 1.0 & 0.2612 & \num{3.12e+03} & \num{.163} & \num{3.32} & \num{1.28} \\
% 1.5 & 0.2515 & \num{2.81e+03} & \num{.143} & \num{2.23} & \num{9.59} \\
% 2.0 & 0.2478 & \num{2.69e+03} & \num{.135} & \num{2.00} & \num{9.01} \\
% \bottomrule
% \end{tabular}
% \end{table*}

\centering
\resizebox{\linewidth}{!}{
\begin{tabular}{c c c c c}
\toprule
\multirow{3}{*}{\textbf{Val CE ($\downarrow$)}} & \multicolumn{4}{c}{\textbf{Test MAE ($\downarrow$)}} \\
\cmidrule(lr){2-5}
 & \textbf{life\_time} & \textbf{cpu\_max} & \textbf{ram\_max} & \textbf{disk\_max} \\
 & \textbf{(min)} & \textbf{(vCPU)} & \textbf{(GB)} & \textbf{(GB)} \\
\midrule
0.2873 & 61.00 & 0.253 & 6.19 & 3.24 \\
0.2612 & 52.00 & 0.163 & 3.32 & 1.28 \\
0.2515 & 46.83 & 0.143 & 2.23 & 0.96 \\
0.2478 & 44.83 & 0.135 & 2.00 & 0.90 \\
\textbf{0.2463} & \textbf{41.07} & \textbf{0.130} & \textbf{1.66} & \textbf{0.76} \\
\bottomrule
\end{tabular}
}
\captionof{table}{Correlation between Validation Loss and Test MAE of Qwen-3-8B on ChipDV-Repetitive. Rows are ordered by training progress.}
\label{tab:ce_mae_correlation}
  \end{minipage}
\end{table*}

\subsection{Ablation Study}\label{subsec:ablation}

\textbf{Correlation between CE Loss and Prediciton Accuracy.}
\ourmethod\ minimizes cross-entropy (CE) loss during fine-tuning rather than a regression objective such as MAE. To verify CE loss is a reliable training signal, we evaluate Qwen-3-8B checkpoints at steps 500, 1000, 1500, 2000, and 2191 (epoch 2) on the ChipDV-Repetitive test set. As shown in \cref{tab:ce_mae_correlation}, validation CE loss and test MAE across all four resource metrics show clear decreasing trend, with the final checkpoint achieving the lowest values on both objectives. This confirms that minimizing CE loss is an effective proxy for regression accuracy in workflow resource prediction.
% \ourmethod's fine-tuning minimizes cross-entropy (CE) loss, rather than an regression metric like MAE. To understand the relationship between two objectives, we tracked both the test CE loss and the test MAE of model checkpoints during fine-tuning on ChipDV-Repetitive. We evaluate intermediate checkpoints of Qwen-3-8B model, from 500, 1000, 1500, 2000, and final steps (2191 at epoch 2) on test data. Results are shown in \cref{tab:ce_mae_correlation} from the first row to the end, respectively. We observe a strong correlation: as the CE loss decreases with more training, the MAE for all resource metrics also consistently decreases. The best performance is achieved at the final training stage, where both CE loss and all MAE values are at their minimum. The results demonstrate that minimizing CE loss is an effective proxy for adapting LLMs on workflow resource prediction task.

\begin{figure*}[!t]
  \centering
  \begin{minipage}[t]{0.49\textwidth}
    \centering
    \includegraphics[width=\linewidth]{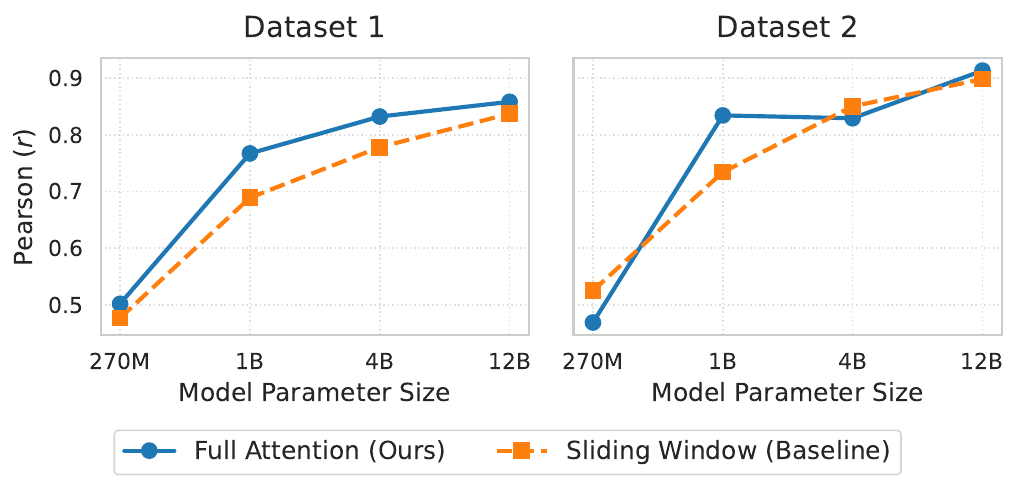}
    \captionof{figure}{Full attention vs.\ sliding window attention (SWA) across Gemma-3 model scales on ChipDV.}
    \label{fig:attention}
  \end{minipage}\hfill
  \begin{minipage}[t]{0.49\textwidth}
    \centering
    \includegraphics[width=\linewidth]{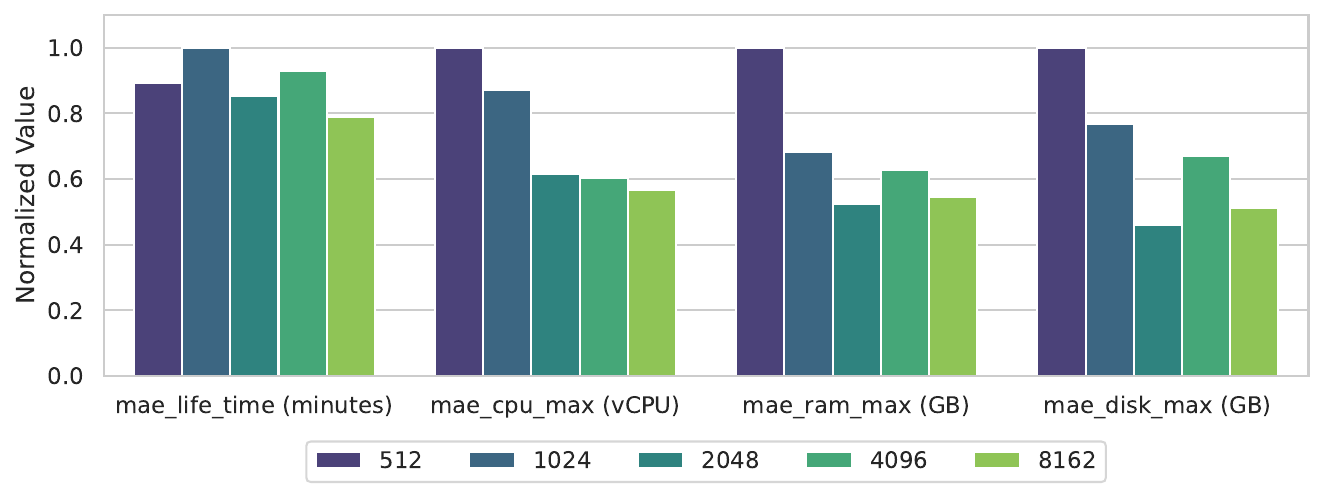}
    \captionof{figure}{Normalized MAE of Gemma-3-4B on ChipDV-Repetitive under varying maximum sequence lengths.}
    \label{fig:ablation_seq_len}
  \end{minipage}
\end{figure*}

\textbf{Efficiency of Constrained Decoding.}
Beyond enforcing valid output format (\cref{fig:system-diagram}), constrained decoding also reduces inference latency. We measure total wall-clock generation time for standard vs.\ constrained decoding on ChipDV-Repetitive across Gemma-3-1B, 4B, and 12B. As shown in \cref{tab:constrained_decoding}, constrained decoding achieves over 30\% latency reduction on all three models, by skipping LLM forward passes for deterministic tokens and reducing the total tokens generated per example.

% In addition to enforcing the correct generation format, whose instances are shown in \cref{fig:system-diagram}, we demonstrate another benefit of the proposed constrained decoding method: temporal efficiency during LLM inference. We compared the total wall-clock time required to generate responses using standard decoding versus our constrained decoding approach. The experiments were conducted on dataset 1 across Gemma-3-1B, Gemma-3-4B, and Gemma-3-12B models.

% As illustrated in \cref{fig:constrained_decoding}, our method significantly reduces the computational cost of generation. Specifically, we observed a latency reduction of more than 30\% for all three models. This speedup is attributed to the constraint mechanism bypassing LLM's forward of deterministic tokens, thereby reducing the total number of tokens generated per example compared to the unconstrained baseline.

\textbf{Full Attention vs.\ Sliding Window Attention.}
Gemma-3 models default to Sliding Window Attention (SWA: 512-token window for 270M/1B, 1024 for 4B/12B). We ablate this by fine-tuning both SWA and full-attention variants on ChipDV-Repetitive and comparing Pearson/Spearman correlation across model scales (\cref{fig:attention}). Full attention consistently matches or outperforms SWA, with the largest gain at 1B. At 12B the gap narrows, but full attention still improves on both datasets. These results indicate that global context visibility is important for resource prediction of SWA LLMs.
% To evaluate the impact of context visibility, we conducted an ablation study using the default Sliding Window Attention (SWA) configurations for Gemma 3 models (512-token sliding window for 270M and 1B, 1024 for 4B and 12B). We fine-tuned them on dataset 1 and compared their Pearson correlation and Spearman correlation to those of the full-attention fine-tuning from \cref{tab:results_combined}.

% As illustrated in \cref{fig:attention}, the Full Attention mechanism (blue line) consistently outperforms or matches the Sliding Window baseline (orange dashed line) across model scales. The benefit of full context is most pronounced in the 1B parameter model, where we observe a performance gap of $+0.078$ on Dataset 1 ($0.767$ vs. $0.689$) and $+0.100$ on Dataset 2 ($0.834$ vs. $0.734$). While the sliding window approach remains competitive at larger scales (e.g., 12B), the full attention mechanism ensures robust performance across both datasets: it improves average $r$ from 0.695 to 0.734 on dataset 1, and from 0.752 to 0.761 on dataset 2, respectively. This enhancement showcases that global context visibility is critical for minimizing regression errors in the EDA job task.

\paragraph{Impact of Maximum Sequence Length} \label{subsec:max-seq-len}
We ablate maximum sequence length on ChipDV-Unseen using Gemma-3-4B, evaluating 512, 1024, 2048, 4096, and 8192 tokens, in \cref{fig:ablation_seq_len}. Longer sequences improve accuracy by retaining more job context, but gains plateau beyond 2048—with 4096 and 8192 providing only marginal improvement. As 2048 also minimizes \texttt{ram\_max} and \texttt{disk\_max} MAE, we adopt it as the default.

% We analyze the maximum sequence length to balance information retention against computational cost. Gemma-3-4B is trained on dataset 1 with different maximum sequence length: 512, 1024, 2048, 4096, and 8192. \cref{fig:ablation_seq_len} shows the normalized MAE across all regression metrics. The result demonstrates that increasing sequence length improves regression accuracy by incorporating more relative job information. However, lengths beyond 2048, like 4096 or 8192, provide marginal benefits. Since 2048 maximum sequence length also minimizes ram and disk prediction error, we adopt it as the optimal setting.

\section{Conclusion}
% We introduced a novel framework for predicting EDA cloud job resource usage and lifetime by leveraging LLMs. By formulating the multi-output regression problem as a sequence-to-sequence generation task, we minimize the need for brittle, manual feature engineering and regression model training, and enable LLMs to learn directly from rich, semi-structured job configuration data. Our primary contributions are the initial application of sequence modeling to this critical industrial setting and the introduction of two key techniques: (1) scientific notation for outputs, and (2) constrained decoding that collectively improve the accuracy, robustness, and numerical stability of the predictions. We also demonstrated the benefit of full attention fine-tuning over a sliding window LLM. The effectiveness of our framework was showcased on challenging real-world EDA datasets, establishing a new and powerful baseline for resource and lifetime prediction.

We presented \ourmethod, a framework for fine-tuning LLMs on semi-structured workflow job configurations for multi-target resource and runtime regression, eliminating the brittle feature engineering that limits traditional ML approaches. Three techniques make this effective: scientific notation encoding for targets spanning multiple orders of magnitude, constrained decoding with prefix filling that enforces output validity while reducing inference latency by over 30\%, and full-attention fine-tuning for long job contexts. Validated on industrial chip design workloads and GHARuntime, a new public benchmark of 580{,}000+ GitHub Actions runs across 27{,}000 repositories, \ourmethod\ consistently outperforms production heuristics and tabular ML baselines, with performance scaling predictably with model size and training data. 

\clearpage

\bibliography{colm2026_conference}

\clearpage
\appendix

% \subsection{Background on EDA Workloads}
% EDA workflows consist of a series of computational jobs, such as logic synthesis, place-and-route, timing analysis, and physical verification. These jobs' %are executed by specialized tools, and their 
% performance is highly dependent on a multitude of factors, including but not limited to:

% \textit{Design Characteristics:} The size and complexity of the circuit design (e.g., number of logic gates, memory blocks).

% \textit{Tool Configuration:} The specific EDA tool, its version, and the multitude of settings and flags used for a particular run.

% \textit{Technology Node:} The target semiconductor manufacturing process (e.g., 7nm, 5nm), which dictates physical design rules.

% \textit{Execution Environment:} The underlying cloud infrastructure, including VM types and storage solutions.

% The interplay between these factors creates a high-dimensional and complex feature space. A minor change in a synthesis script could lead to a drastically different netlist and cause a tenfold increase in the runtime of the subsequent place-and-route stage. This sensitivity makes prediction a complicated regression task.

\section{Background on Chip Design and CI/CD Workflow Workloads}
\label{sec:background}

Modern cloud workflow jobs span domains from chip design to continuous integration and continuous deployment~(CI/CD). Despite their surface differences, both share a defining structure: heterogeneous, semi-structured job configurations whose runtime and resource demands are difficult to predict from numerical metadata alone~\citep{liu2023cost, bavikadi2022survey}.

\paragraph{Chip Design Workloads.}
Chip design workflows, formally electronic desgin automation (EDA) workflows, consist of a series of computational jobs such as logic synthesis, place-and-route, timing analysis, and physical verification~\citep{stok2014eda}. As semiconductor process technology advances, even a single sub-5-million-gate partition can require over 25 days of wall-clock time to complete a full EDA flow at the 3nm node~\citep{jiang2024eda}, making accurate runtime and resource prediction essential for project scheduling and cloud cost control~\citep{zhu2024elastic}. Our industrial dataset focuses specifically on \textbf{Design Verification~(DV)} workloads \citep{stok2014eda}, where jobs exercise a chip's functional correctness across combinatorial-scale simulation runs. The performance of DV jobs depends critically on four classes of factors. \textit{Design Characteristics} encompass the size and complexity of the circuit under test, including the number of logic gates, memory blocks, and testbench stimulus files, all of which directly govern simulation time. \textit{Tool Configuration} covers the specific EDA simulator, its version, and the multitude of settings and flags---such as coverage collection modes and compile-time optimizations---that can cause order-of-magnitude runtime differences between otherwise similar jobs~\citep{huang2021eda}. \textit{Technology Node} refers to the target semiconductor process (e.g., 7nm, 5nm), which determines physical design rules and indirectly influences simulation model complexity. \textit{Execution Environment} includes the underlying cloud VM type, storage tier, and license availability, all of which introduce significant runtime variability even for identical job configurations~\citep{bavikadi2022survey}. The interplay between these factors creates a high-dimensional, semi-structured feature space that resists conventional tabular encoding~\citep{jiang2024eda, zhu2024elastic}.

\paragraph{CI/CD Workloads.}
CI/CD workflow jobs, as captured in our \textbf{GHARuntime} benchmark, are triggered by repository events~(e.g., \texttt{push}, \texttt{pull\_request}, \texttt{schedule}) and execute on cloud-hosted virtual machines~\citep{moriconi2025ghalogs}. GitHub Actions has emerged as the dominant CI/CD platform, with prior work showing that testing and building together consume around 90\% of total CI/CD VM time~\citep{bouzenia2024gha}. Each job is a sequence of heterogeneous steps---GitHub Actions with tool-specific parameters~(e.g., \texttt{actions/setup-python} with \texttt{python-version:3.9}) and shell commands~(e.g., \texttt{cibuildwheel} targeting \texttt{aarch64}) --- whose runtimes span three orders of magnitude, from sub-minute utility checks to multi-hour cross-compilation builds. 

Researchers recently observe that predicting a near-optimal timeout value for a CI/CD job requires reasoning about the job configuration itself and its repository history ~\citep{bouzenia2024gha}, a signal that purely numerical schedulers cannot capture. Both chip design and CI/CD workloads share the defining challenge: the most predictive signals reside in \emph{textual, semi-structured} configuration fields that tabular feature engineering cannot faithfully encode~\citep{liu2025ora, huang2021eda}.

\section{Experimental Setup Details}\label{app:exp-setup}

\subsection{GHARuntime System Prompt}

The system prompt used for \ourmethod\ on GHARuntime is as follows:

\begin{tcolorbox}[
    colback=gray!5,
    colframe=gray!40,
    boxrule=0.4pt,
    arc=3pt,
    left=8pt, right=8pt, top=6pt, bottom=6pt,
    fontupper=\small\ttfamily,
    title={\small\sffamily\bfseries System Prompt --- GHARuntime},
    coltitle=black,
    attach boxed title to top left={yshift=-2mm, xshift=4mm},
    boxed title style={colback=white, colframe=gray!40, boxrule=0.4pt, arc=2pt}
]
You are a precise CI/CD workflow job runtime forecaster.\\[4pt]
INPUT: A JSON object describing a workflow job, including runner image, action steps, and shell commands.\\[4pt]
TASK: Predict the actual wall-clock runtime in seconds.\\[4pt]
OUTPUT: Exactly one JSON object with no surrounding prose:\\
\{"duration\_sec": "<value>"\}\\[4pt]
NUMERIC FORMAT: Python f"\{x:.2e\}" notation.\\
- Exactly 2 fractional digits, e.g. 12300 $\to$ "1.23e+04"\\
- No leading + on mantissa, exponent always has sign.
\end{tcolorbox}

\subsection{GHARuntime Input Example}

Below is a representative GHARuntime job record as consumed by \ourmethod. The job belongs to the \texttt{selinuxproject/refpolicy} repository, triggered by a pull request on an \texttt{ubuntu-20.04} runner, and has a ground-truth runtime of $206.0$s (encoded as \texttt{"2.06e+02"}).

\begin{tcolorbox}[
    colback=blue!2,
    colframe=gray!40,
    boxrule=0.4pt,
    arc=3pt,
    left=8pt, right=8pt, top=6pt, bottom=6pt,
    fontupper=\footnotesize\ttfamily,
    title={\small\sffamily\bfseries LLM Input --- Serialized Job Configuration},
    coltitle=black,
    attach boxed title to top left={yshift=-2mm, xshift=4mm},
    boxed title style={colback=white, colframe=gray!40, boxrule=0.4pt, arc=2pt}
]
\{\\
\hspace*{1em}"repo": "selinuxproject/refpolicy",\\
\hspace*{1em}"event": "pull\_request",\\
\hspace*{1em}"image": "ubuntu-20.04",\\
\hspace*{1em}"num\_steps": 8,\ "num\_shell": 6,\ "num\_action": 2,\\
\hspace*{1em}"steps": [\\
\hspace*{2em}\{"type": "action", "action": "actions/checkout",\\
\hspace*{3em}"with": \{"fetch-depth": "1", "lfs": "false", ...\}\},\\
\hspace*{2em}\{"type": "action", "action": "actions/setup-python",\\
\hspace*{3em}"with": \{"python-version": "3.5", ...\}\},\\
\hspace*{2em}\{"type": "shell", "code": "sudo apt-get install -qy bison flex ..."\},\\
\hspace*{2em}\{"type": "shell", "code": "make bare \&\& make conf \&\& make ..."\},\\
\hspace*{2em}...\\
\hspace*{1em}]\\
\}
\end{tcolorbox}

\begin{tcolorbox}[
    colback=green!3,
    colframe=gray!40,
    boxrule=0.4pt,
    arc=3pt,
    left=8pt, right=8pt, top=6pt, bottom=6pt,
    fontupper=\small\ttfamily,
    title={\small\sffamily\bfseries LLM Output --- Predicted Target},
    coltitle=black,
    attach boxed title to top left={yshift=-2mm, xshift=4mm},
    boxed title style={colback=white, colframe=gray!40, boxrule=0.4pt, arc=2pt}
]
\{"duration\_sec": "2.06e+02"\}
\end{tcolorbox}

\subsection{Tabular Feature Representation}

For tabular ML baselines (XGBoost, LightGBM, and TabPFN), the rich semi-structured job configuration is flattened into a fixed-size feature vector. This process discards hierarchical structure and shell command semantics, retaining only hand-engineered scalar indicators. The feature set for the same example is shown below.

\begin{tcolorbox}[
    colback=orange!3,
    colframe=gray!40,
    boxrule=0.4pt,
    arc=3pt,
    left=8pt, right=8pt, top=6pt, bottom=6pt,
    fontupper=\footnotesize\ttfamily,
    title={\small\sffamily\bfseries Tabular Baseline --- Flattened Feature Vector},
    coltitle=black,
    attach boxed title to top left={yshift=-2mm, xshift=4mm},
    boxed title style={colback=white, colframe=gray!40, boxrule=0.4pt, arc=2pt}
]
num\_steps: 8,\ num\_shell: 6,\ num\_action: 2,\ total\_code\_len: 1079\\
is\_ubuntu: 1,\ is\_macos: 0,\ is\_windows: 0\\
is\_push: 0,\ is\_pr: 1,\ is\_schedule: 0\\
act\_checkout: 1,\ act\_setup-python: 1,\ act\_setup-node: 0,\ ...\\
shell\_make: 1,\ shell\_pytest: 0,\ shell\_npm\ test: 0,\ ...\\
has\_python\_ver: 0,\ has\_node\_ver: 0,\ has\_aarch64: 0
\end{tcolorbox}

\noindent This comparison illustrates the information loss inherent in tabular feature engineering: shell command content, action parameter values (e.g., \texttt{python-version: 3.5}), and step ordering are all discarded, while \ourmethod\ consumes the full structured configuration directly.

\section{GHARuntime: Dataset Construction}
\label{app:gharuntime}

\subsection{Source and Motivation}
\label{app:gharuntime-source}

Public workflow datasets such as WfCommons~\citep{coleman2022wfcommons}
predominantly contain homogeneous task structures in which the same
binary is applied to different input partitions, providing limited
semantic diversity for LLM-based prediction methods.
To address this gap, we construct \textbf{GHARuntime}, a large-scale
public benchmark for workflow job runtime prediction derived from the
GHALogs corpus~\citep{moriconi2025ghalogs}.
GHALogs contains over 116{,}000 CI/CD workflows executed via GitHub
Actions~(GHA) collected from more than 25{,}000 public repositories
across 20 programming languages, comprising approximately 513{,}000
workflow runs and 2.3~million individual steps.
Unlike scientific workflow collections, GitHub Actions workflows exhibit
\emph{genuinely heterogeneous} job configurations: each job is a
sequence of GitHub Actions and shell steps whose tool invocations,
parameters, and runtime characteristics vary widely across repositories
and programming ecosystems.

\subsection{Extraction Pipeline}
\label{app:gharuntime-pipeline}

We process the GHALogs \texttt{runs.json} file
(580{,}641~workflow run records) through the following steps:

\paragraph{Step 1: Filtering.}
We retain only workflow runs with \texttt{conclusion == "success"}
to ensure that all runtime labels reflect complete, valid executions.
Runs with missing or malformed \texttt{log\_insights} fields are
discarded.

\paragraph{Step 2: Job extraction.}
Each workflow run contains a \texttt{log\_insights} list in which
each entry corresponds to one CI/CD job (a parallel unit of
execution assigned to a dedicated virtual machine).
For each job, we sum the \texttt{duration\_sec} field of all
constituent steps to obtain the job-level wall-clock duration:
\begin{equation}
  d_{\text{job}} = \sum_{s \in \text{steps}(j)} d_s.
  \label{eq:job-duration}
\end{equation}
Jobs with $d_{\text{job}} \leq 0$ or no steps are discarded.

\paragraph{Step 3: Serialization.}
Each job is serialized into a structured JSON object containing:
\begin{itemize}[leftmargin=*]
  \item \texttt{event}: the GitHub event that triggered the run
        (e.g., \texttt{push}, \texttt{pull\_request},
        \texttt{schedule});
  \item \texttt{image}: the runner virtual machine image
        (e.g., \texttt{ubuntu-22.04}, \texttt{macos-12},
        \texttt{windows-2022});
  \item \texttt{steps}: an ordered list of step descriptors.
        For \texttt{action}-type steps, each descriptor includes
        the action name (e.g.,
        \texttt{actions/setup-python}) and its \texttt{with}
        parameters (e.g., \texttt{python-version: "3.11"}).
        For \texttt{shell}-type steps, each descriptor includes
        the full shell command string (\texttt{code} field),
        truncated to 300~characters.
        Sensitive fields such as authentication tokens are
        filtered before serialization.
\end{itemize}
This serialization mirrors the input format used for the chip
design simulation datasets, enabling the same \textsc{LASER}
pipeline to be applied without modification.

\paragraph{Step 4: Train/test split.}
To ensure that test repositories are completely unseen during
training---a stricter evaluation protocol than random
splitting---we partition by \emph{repository}:
all jobs from 80\% of repositories form the training set and
all jobs from the remaining 20\% form the test set.
The split is deterministic (fixed random seed~42 after shuffling
the sorted repository list).

\subsection{Dataset Statistics}
\label{app:gharuntime-stats}

\begin{table}[h]
  \centering
  \caption{GHARuntime dataset statistics.}
  \label{tab:gharuntime-stats}
  \begin{tabular}{lr}
    \toprule
    \textbf{Statistic} & \textbf{Value} \\
    \midrule
    Total workflow runs (source)    & 580{,}641  \\
    Total workflow runs (after filtering failed runs) & 440{,}109 \\
    Total jobs (after filtering)    & 1{,}322{,}504 \\
    Unique repositories             & 27{,}728   \\
    Unique workflow paths           & 33{,}310   \\
    Median job duration             & 100.3\,s   \\
    90th-percentile duration        & 809.3\,s   \\
    99th-percentile duration        & 3{,}865.7\,s \\
    Maximum duration                & 108{,}557.3\,s \\
    Std / Mean ratio (CoV)          & 2.80       \\
    Training jobs                   & $\sim$1.06M \\
    Test jobs                       & $\sim$0.26M \\
    \bottomrule
  \end{tabular}
\end{table}

The duration distribution is heavily right-skewed (coefficient of
variation~2.80), spanning three orders of magnitude from
sub-minute utility jobs (e.g., \texttt{twine check}, 12\,s) to
multi-hour cross-compilation builds
(e.g., \texttt{aarch64} wheel builds exceeding 4\,hours).
This range and skew closely mirror the distribution observed in
the chip design simulation datasets (Figure~2 of the main paper),
making GHARuntime a natural out-of-domain complement for
evaluating \textsc{LASER}'s generalization.

\begin{figure}[t]
  \centering
  \includegraphics[width=\linewidth]{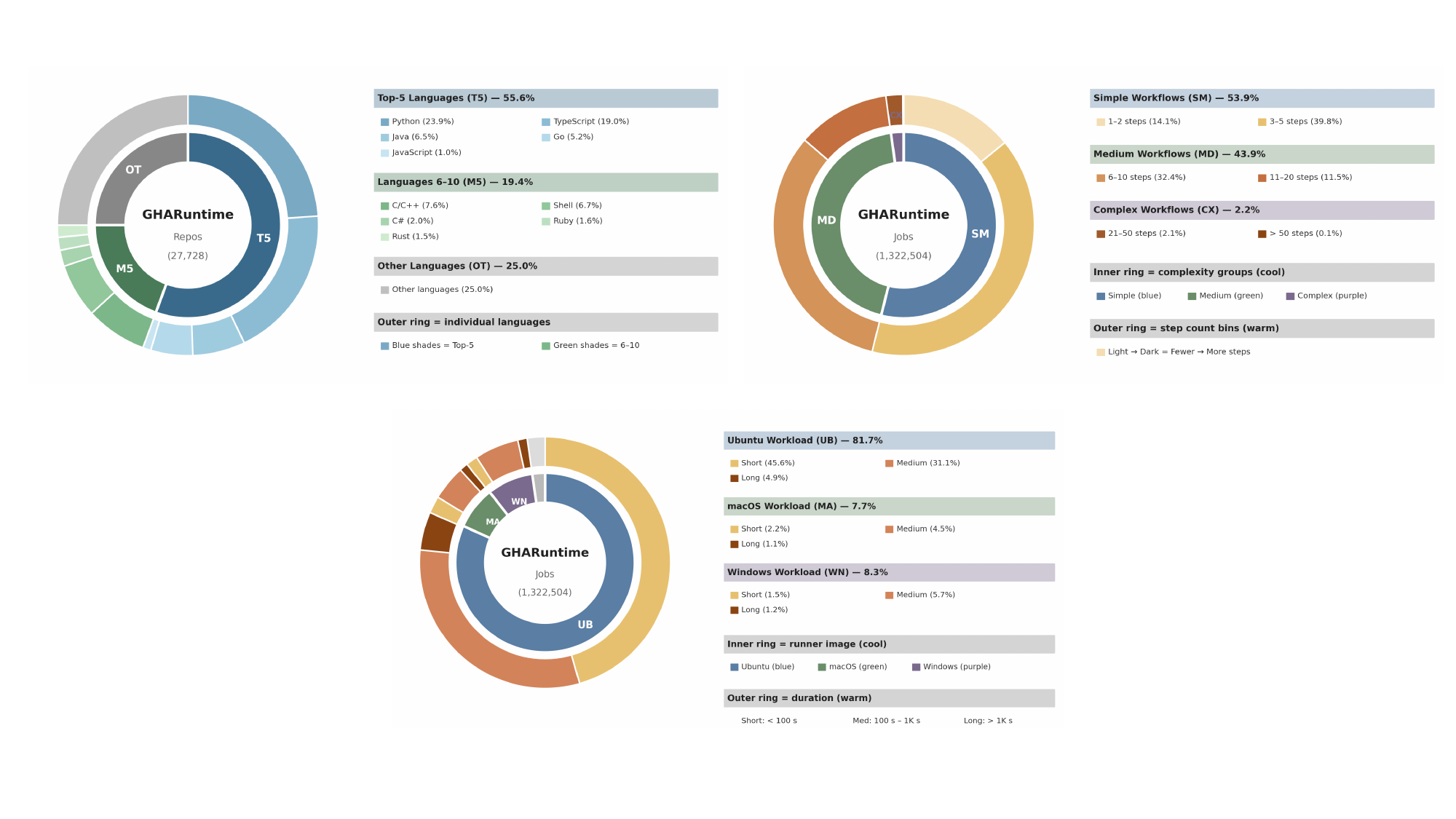}
  \caption{Distribution statistics of GHARuntime job durations on
  training and test splits. The figure highlights the long-tail
  structure and split-level consistency.}
  \label{fig:gharuntime-dataset-stats}
\end{figure}
Figure~\ref{fig:gharuntime-dataset-stats} summarizes the composition of GHARuntime across three dimensions.
\textbf{By runner image}, Ubuntu dominates at 81.7\% of all jobs, followed by Windows (8.3\%) and macOS (7.7\%), reflecting the prevalence of Linux-based CI/CD workflows on GitHub Actions.
Within each platform, the duration breakdown reveals that short jobs ($<$100\,s) constitute the majority of Ubuntu workloads (45.6\%), while macOS and Windows jobs skew toward medium durations (100\,s--1\,Ks), consistent with their heavier use in build and compilation tasks.
\textbf{By programming language}, the dataset spans 20 languages across 27,728 repositories, with Python (23.9\%) and TypeScript (19.0\%) as the two most represented.
The top-5 languages account for 55.6\% of repositories, while a long tail of 10+ additional languages contributes 25.0\%, ensuring semantic diversity in workflow configurations.
\textbf{By workflow complexity}, over half of all jobs (53.9\%) involve simple workflows with 1--5 steps, while 43.9\% fall in the medium range (6--20 steps).
Only 2.2\% of jobs exceed 20 steps, representing complex multi-stage pipelines such as matrix builds and cross-platform release workflows.

\subsection{Why GHARuntime Enables LLM-Based Prediction}
\label{app:gharuntime-why}

The key property distinguishing GHARuntime from prior workflow
trace collections is the richness of \emph{command-level text
metadata} available at prediction time.
Consider two jobs with identical structural metadata
(same runner image, same number of steps):
one runs \texttt{python -m tables.tests.test\_all} (225\,s)
and another runs a \texttt{cibuildwheel} cross-compilation
targeting \texttt{aarch64} (15{,}952\,s)---a 70$\times$
difference invisible to any tabular feature.
\textsc{LASER} encodes the full shell command and action
parameter strings, allowing it to capture such signals directly.
In contrast, tabular ML baselines can only observe keyword
indicators hand-engineered from the command text, necessarily
losing precision on long-tail tool invocations.

\end{document}